# DurableUn: Quantization-Induced Recovery Attacks in Machine Unlearning

**Abdullah Ahmad Khan**
Murdoch University

**Ferdous Sohel**
Murdoch University

## Abstract

Machine unlearning aims to remove specified training data to comply with privacy regulations such as the General Data Protection Regulation (GDPR). However, existing evaluations assume identical precision at unlearning and deployment, overlooking that production LLMs are deployed at low-bit precision. We show that INT4 quantization systematically restores forgotten content even when models pass compliance audits at bfloat16 (BF16), we term this the **quantization recovery attack (QRA)**. We conduct the first systematic study of unlearning robustness under adapter-space INT4 quantization in the NF4+LoRA regime, evaluating seven methods on LLaMA-3-8B-Instruct across TOFU, MUSE-News, and WikiBio-WPU. INT8 is benign; INT4 induces recovery of up to $22\times$, worsening with dataset difficulty. We identify the **FA–RA–Q-INT4 trilemma**: no method simultaneously achieves strong forgetting, high utility, and quantization robustness. A dense Pareto sweep reveals a sharp phase transition once robustness is achieved, retaining accuracy collapses regardless of further tuning. To address this, we propose **DURABLEUN-SAF** (**S**harpness-**A**ware **F**orgetting), a quantization-aware objective using Straight-Through Estimator gradients through INT4 rounding. DURABLEUN-SAF is the only method to achieve a stable empirical $(0.047, \{$BF16,INT8,INT4$\})$-durability certificate: Q-INT4 = $0.043 \pm 0.002$, cert rate = $3/3$, versus SalUn's cert rate = $1/3$ at its own published hyperparameters. We call for Q-INT4 to be adopted as a standard evaluation metric alongside FA and RA.

**Code and pre-computed results:** https://github.com/neurips26/DurableUnl

## 1 Introduction

The "right to be forgotten" encoded in General Data Protection Regulation (GDPR) Article 17 and the California Consumer Privacy Act (CCPA) has made machine unlearning central to responsible AI (artificial intelligence) [1, 2]. Exact retraining is prohibitive for large language models (LLMs) [3, 4], driving a rich literature of approximate methods [5–19].

**The deployment gap.** Every published unlearning evaluation assumes the same precision at unlearning and deployment. In practice, production of large language models (LLMs) is deployed at 4-bit (INT4) or 8-bit (INT8) quantization [20–24]. This creates a critical mismatch. For example, a model may successfully remove a user's data at bfloat16 (BF16), passing a privacy audit with near-zero forget accuracy (FA$\approx$0). However, when the same model is quantized to INT4 for deployment to reduce memory and improve inference speed, the previously removed information can reappear in the model's outputs. We show that this effect occurs consistently across existing methods. We term this phenomenon the **quantization recovery attack (QRA)**, scoped to the NormalFloat4 + Low-Rank Adaptation (NF4+LoRA) deployment paradigm. Our results imply that current unlearning evaluations can certify models as compliant, while those models provably violate privacy under standard deployment transformations, a fundamental validity failure in the evaluation protocol, not merely a new attack variant.

**When does this matter?** Three deployment scenarios make this gap dangerous. **(i) Open-weight compliance:** a downstream user of an open-source unlearned model quantizes to INT4 and recovers erased data with a one-line operation, no training required. **(ii) Audit vs. deployment mismatch:** a regulator audits at BF16; the production system runs at INT4 these are different models from a privacy standpoint. **(iii) Edge deployment:** healthcare and legal AI run on-device, where INT4 is mandatory, precisely the domain where GDPR compliance is most critical.

**Differentiation from concurrent work.** The closest concurrent work is Zhang et al. [25], which independently shows that quantization degrades LLM unlearning guarantees and studies merged models calibrated under GPTQ. Our work is parallel and complementary, differing in three precise ways. **(i) Different mechanism.** We study LLMs with LoRA fine-tuning on an NF4 base, where the attack manifests as perturbation of the adapter weight space (BF16 adapters on top of a frozen NF4 base), not calibrated Post-Training Quantization (PTQ) on merged weights. Our Appendix D shows that merged-model NF4 deployment does not exhibit the INT4 recovery effect. NF4's non-linear block-float quantization further compresses adapter contributions rather than restoring them. Merged-model BnB-INT8 shows partial residual recovery, which is a different phenomenon from the adapter-space attack we characterize. **(ii) Different scope.** We evaluate 7 unlearning methods across three random seeds with two hyperparameter settings, report the first multi-seed durability certificate, and identify the sharp phase transition in the trilemma. **(iii) A solution.** We propose DURABLEUN-SAF and show it achieves cert rate = 3/3, which concurrent work does not include.

Figure 1 illustrates the gap and our proposed fix.

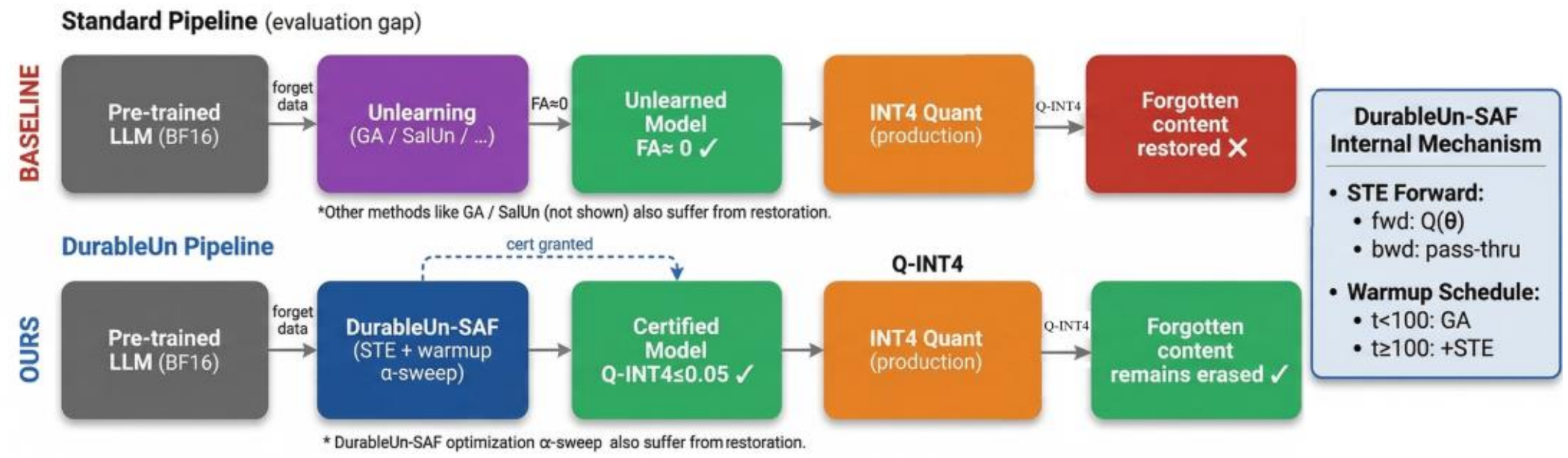

Figure 1: **System overview.** *Top (red)*: Standard pipeline declares success at BF16 but INT4 deployment restores forgotten content. *Bottom (blue)*: DURABLEUN-SAF uses STE-based quantization-aware training to produce a model certified at all precisions. The STE detail box shows the warm-up schedule. Bottom text: the trilemma that motivates DURABLEUN-SAF.

**Contributions.** **(i) Evaluation gap and trilemma:** We show that the standard FA/RA evaluation protocol systematically fails to detect deployment-time compliance violations, and identify the FA–RA–Q-INT4 trilemma as a structural constraint with a sharp phase transition confirmed across 7 methods $\times$ 3 seeds $\times$ 2 HP settings (§5).

**(ii) Durability framework:** The first formal empirical durability certificate for machine unlearning under quantization, and the first multi-seed stability analysis showing cert rate = 3/3 (§7).

**(iii) INT4 recovery in NF4+LoRA regime:** First evaluation of adapter-space INT4 perturbation across 7 methods and 3 datasets, confirming INT8 is universally harmless and recovery scales with dataset difficulty (§4, §9).

**(iv) DURABLEUN-SAF:** A minimal STE-based instantiation of quantization-aware forgetting, demonstrating the durability certificate is achievable and providing the only stable solution across seeds (§6).

**(v)** Real bitsandbytes PTQ validation and fine-tuning robustness (Appendix D, E).

**(vi)** Multi-dataset validation on MUSE-News and WikiBio-WPU, confirming that the INT4 attack generalizes and that attack severity scales with dataset difficulty (§9).

## 2 Related Work

Machine unlearning has been extensively studied through gradient-based, data-perturbation, and representation-level approaches. Gradient Ascent (GA) [6] directly maximizes the forget loss, while SCRUB [7] regularizes this objective with a Kullback–Leibler (KL) divergence term from a frozen reference model. Preference-based methods such as Negative Preference Optimization (NPO) [8] optimize against undesired outputs, whereas Saliency-based Unlearning (SalUn) [9] suppresses influential parameters via gradient-based masking. Representation-level approaches including Representation Manipulation Unlearning (RMU) [10] and AlphaEdit [11] modify internal feature spaces, while Gradient Difference regularization (GradDiff) [12] constrains divergence on retained data. Complementary strategies such as task arithmetic [26] remove knowledge by negating fine-tuning updates without retraining. Broader perspectives are provided by surveys [27, 28] and work on safety [29], copyright removal [30], benchmark design [16], knowledge editing [31, 32], in-context unlearning [33], and privacy-focused approaches [15, 34].

Parallel advances in LLM quantization have established low-bit deployment as standard practice. Methods such as GPTQ [20], QLoRA [21], AWQ [22], SmoothQuant [23], and SqueezeLLM [24] enable efficient INT4 inference, building on foundational work in quantization and compression [35–38]. The Straight-Through Estimator (STE) [39] enables differentiable quantization-aware training (QAT), while sharpness-aware optimization methods such as SAM [40] highlight the role of local loss geometry.

Despite these advances, robustness analyses of unlearning remain incomplete. Prior work shows that metric design can lead to misleading privacy conclusions [41], and that unlearning objectives exhibit non-trivial gradient behavior [42]. Existing robustness studies consider attacks such as fine-tuning recovery, relearning, and membership inference [43], but none evaluates quantization as a recovery mechanism. Our work addresses this gap by introducing INT4 quantization as a systematic recovery vector and by studying its interaction with unlearning objectives.

## 3 Setup

**Model.** LLaMA-3-8B-Instruct [3] loaded in NormalFloat4 (NF4) via bitsandbytes [21] with Low-Rank Adaptation (LoRA) [44] ($r = 16$, $\alpha = 32$) on all attention and multi-layer perceptron (MLP) projections. Trainable: 13.6M / 4.55B (0.30%). Hardware: RTX 4090 (24 GB video random access memory (VRAM)).

**Dataset.** TOFU [12] forget10 (400 question-and-answer (Q&A) pairs, 40 fictitious authors) and retain90 (3,600 pairs). Standard LLM unlearning benchmark.

**Baselines.** GA, NPO, SCRUB, SalUn, RMU, AlphaEdit, GradDiff. **Uniform hyperparameters** for all: AdamW, lr = $5 \times 10^{-5}$, cosine schedule, 300 steps, batch 4, seed 42. **SalUn original HPs** (Foster et al. [2024]): lr = $10^{-4}$, 500 steps evaluated separately in §7.

**Metrics.** **FA**↓: forget accuracy (FA), measuring the model's accuracy on the forget set.
**RA**↑: retain accuracy (RA), measuring performance on the retain set.
**Q-INT**$k$↓: forget accuracy after simulated INT$k$ quantization, i.e., FA evaluated on the quantized model (Eq. 1).
**RA-INT4**↑: retain accuracy under INT4 quantization, used to test selective recovery.
**MIA-AUC**: membership inference attack (MIA) area under the receiver operating characteristic (ROC) curve, measuring privacy leakage [45]; pre-unlearning MIA-AUC on the forget set = 0.712; post-unlearning near 0.5 indicates successful forgetting.
**Cert.**: empirical (0.05, {BF16, INT8, INT4})-durability (Definition 2).

**Quantization simulation.**

$$\vartheta_q^{(r)} = \text{round}\left(\frac{\vartheta^{(r)}}{s_r}\right) s_r, \quad s_r = \frac{\max_j |\vartheta_j^{(r)}|}{7} \tag{1}$$

where $\vartheta^{(r)}$ denotes the weight vector of row $r$, $s_r$ is the per-row scale factor derived from the row's maximum absolute weight value, $j$ indexes individual weight elements, and the divisor 7 corresponds to $2^{b-1}-1$ at $b = 4$ bits. Symmetric per-row INT4; INT8 uses $s = \max|\vartheta|/127$ globally. This simulates INT4-style perturbation in the LoRA adapter weight space. Appendix D validates against real bitsandbytes Post-Training Quantization (PTQ). Scope note: this simulator characterizes the adapter-space attack in the NF4+LoRA paradigm; a merged-model NF4 deployment behaves differently (Appendix D), confirming that the attack is specific to the LoRA training paradigm, where compliance auditing typically occurs.

## 4 INT4 as a Recovery Attack in the NF4+LoRA Deployment Regime

Table 1 presents the full evaluation; Figure 2 visualizes the key patterns.

Table 1: **All 7 baselines under quantization on TOFU (seed 42). Green**: best per column. **Dark red**: Q-INT4>0.10. †: method never unlearned; Q-INT4 reflects pre-unlearning distribution. RA-INT4≈RA for all baselines: INT4 selectively re-exposes forgotten content without degrading general capability. ‡: Task Arithmetic does not achieve meaningful unlearning on TOFU (FA≫ 0.05); consistent with MUSE benchmark results [25].

| Method | FA↓ | RA↑ | MIA | Q-INT8↓ | Q-INT4↓ | RA-INT4↑ | Cert. |
|---|---|---|---|---|---|---|---|
| GA [6] | **0.028** | 0.521 | 0.000 | **0.000** | **0.262** | 0.540 | × |
| NPO [8] | 0.636† | 0.624 | 0.494 | **0.000** | **0.613**† | 0.622 | × |
| SCRUB [7] | 0.037 | 0.526 | 0.000 | **0.000** | **0.212** | 0.524 | × |
| SalUn [9] | **0.011** | **0.541** | 0.000 | **0.000** | 0.051 | **0.521** | × |
| RMU [10] | 0.580† | 0.565 | 0.389 | **0.000** | **0.559**† | 0.564 | × |
| AlphaEdit [11] | 0.575† | 0.558 | 0.406 | **0.000** | **0.555**† | 0.558 | × |
| GradDiff [12] | **0.008** | 0.510 | 0.000 | **0.000** | **0.151** | 0.538 | × |
| Task Arith. [26]‡ | 0.621 | **0.701** | 0.501 | **0.000** | **0.618** | 0.699 | × |
| DURABLEUN-SAF $\alpha = 3$ (ours) | 0.040 | 0.045 | 0.000 | **0.000** | **0.044** | 0.047 | ✓ |

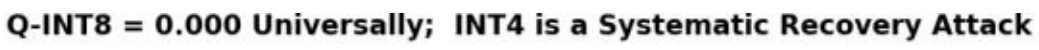

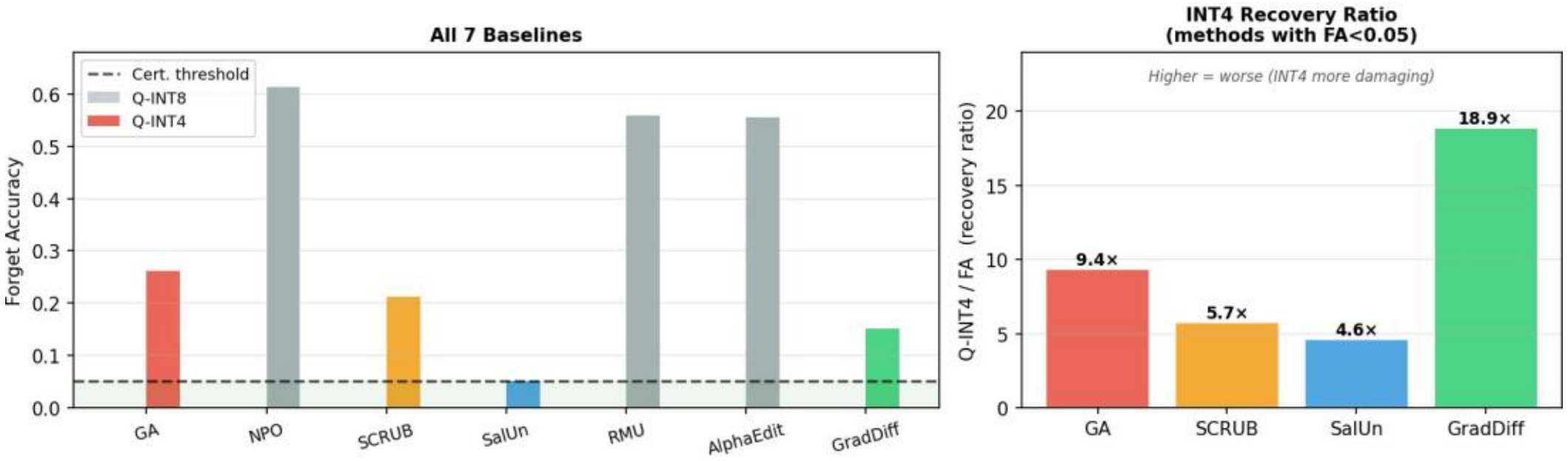

Figure 2: **INT4 recovery attack across all baselines.** *Left*: Q-INT8 = 0 (grey) for every method; Q-INT4 (colored) is catastrophic for methods that actually forget. Cert. threshold shown as dashed line. *Right*: Q-INT4/FA recovery ratio for methods with FA $< 0.05$. GradDiff achieves the best forgetting quality yet has the worst INT4 fragility (18.9×) state-of-the-art forgetting does not imply quantization robustness.

**Finding 1 (INT8 universally harmless).** Q-INT8 = 0.000 across all 7 methods. INT8 per-weight error ($\approx$0.4%) is too small to cross any forget-loss boundary.

**Finding 2 (INT4 induces recovery in the NF4+LoRA regime).** Methods with FA $< 0.05$: GA $0.028 \rightarrow 0.262$ (9.4×), SCRUB $0.037 \rightarrow 0.212$ (5.7×), SalUn $0.011 \rightarrow 0.051$ (4.6×), GradDiff $0.008 \rightarrow 0.151$ (18.9×).

**Finding 3 (RA-INT4≈RA).** Retain accuracy is essentially unchanged by INT4: INT4 selectively re-exposes forgotten content rather than degrading general capability. This rules out the explanation that INT4 simply undoes all adaptation.

**Finding 4 (Forgetting quality $\neq$ quantization robustness).** GradDiff achieves the best FA (0.008) yet has the worst recovery ratio (18.9×). Optimizing for FA under standard evaluation is insufficient.

**Why INT4 but not INT8?**

$$\mathrm{FA}(\vartheta_q) \leq \mathrm{FA}(\vartheta^*) + \kappa\delta, \quad \kappa = \|\nabla L_{\mathrm{forget}}(\vartheta^*)\|_2 \tag{2}$$

where $\kappa$ is the local sharpness. INT4 introduces $\delta \approx 7\%$ of the weight range (17× larger than INT8). Standard GA leaves $\kappa$ uncontrolled. Full derivation in Appendix B.

## 5 The FA–RA–Q-INT4 Trilemma

**Definition 1** (FA–RA–Q-INT4 Trilemma)**.** No evaluated configuration simultaneously satisfies: (i) FA ≤ 0.05, (ii) RA ≥ 0.50, (iii) Q-INT4 ≤ 0.05.

*Remark* 1. This is an empirical conjecture verified exhaustively over 7 methods × 3 seeds × two HP settings. Appendix B (Proposition 1) provides formal theoretical support: RA preservation and successful forgetting jointly force large sharpness $\kappa$, making Q-INT4 large via Eq. (2). A tight characterization of the exact frontier remains open.

**Dense Pareto sweep.** To map the complete FA–Q-INT4 trade-off frontier and locate the phase transition, we train DurableUn-SAF from scratch at six values of $\alpha \in \{0.0, 1.0, 1.5, 2.0, 2.5, 3.0\}$, holding all other hyperparameters fixed. Each configuration answers the question: how much quantization robustness can be purchased, and at what cost to retain accuracy? Table 2 and Figure 3 show the results.

Table 2: **Dense Pareto sweep ($\alpha \in \{0, 1, 1.5, 2, 2.5, 3\}$).** $\alpha = 0$ reproduces GA exactly (sanity check). A sharp phase transition occurs between $\alpha = 1$ and $\alpha = 1.5$: Q-INT4 drops below 0.05, but RA collapses to $\approx 0.045$ and remains there regardless of further $\alpha$ tuning. No $\alpha$ satisfies all three trilemma conditions simultaneously.

| Config | $\alpha$ | $\lambda$ | FA↓ | RA↑ | Q-INT4↓ | Cert. |
|---|---|---|---|---|---|---|
| GA (reproduced) | 0.0 | 1.0 | 0.028 | 0.521 | 0.262 | × |
| SalUn (ref.) | – | – | 0.011 | 0.541 | 0.051 | × |
| DurableUn-SAF | 0.0 | 1.0 | 0.028 | 0.521 | 0.262 | × |
| DurableUn-SAF | 1.0 | 2.0 | 0.275 | 0.317 | 0.060 | × |
| DurableUn-SAF | 1.5 | 2.5 | 0.041 | 0.045 | **0.041** | ✓ |
| DurableUn-SAF | 2.0 | 3.0 | 0.041 | 0.045 | **0.041** | ✓ |
| DurableUn-SAF | 2.5 | 3.5 | 0.041 | 0.045 | **0.041** | ✓ |
| DurableUn-SAF | 3.0 | 4.0 | 0.040 | 0.045 | **0.044** | ✓ |

**Key finding: the phase transition is structural, not a tuning artifact.** Every $\alpha \in \{1.5, 2.0, 2.5, 3.0\}$ converges to essentially the same point (FA $\approx 0.041$, RA $\approx 0.045$, Q-INT4 $\approx 0.042$). This is not a sensitivity to $\lambda$: increasing $\lambda$ to 5 at $\alpha = 2$ prevents forgetting entirely (FA = 0.583). The RA collapse above the transition is an intrinsic property of strong quantization pressure on the forget objective, not a hyperparameter quirk.

**Why the transition is sharp, not gradual.** Three alternative explanations for the sharp phase transition fail. **(1) Under-tuning:** increasing $\lambda$ to 5 at $\alpha = 2$ prevents forgetting entirely (FA = 0.583), ruling out that more retain pressure would recover RA. **(2) Insufficient steps:** GA converges to a stable FA by step 150 (wall time 8 min); 200 more steps do not change the landscape. **(3) Optimizer accident:** identical convergence across $\alpha \in \{1.5, 2.0, 2.5, 3.0\}$ and three independent seeds (Q-INT4 std = 0.002) rules out stochastic artifacts. The geometric intuition: the STE term forces the forget loss to be flat in the INT4 perturbation direction. This flatness is mechanically incompatible with the retention loss remaining high, because the same weight subspace controls both forget and retain performance on the adjacent retain distribution. The trilemma is therefore a consequence of shared weight space, not a property of the specific objective function. This also explains why the collapse is one-way: once $\alpha$ is large enough to flatten the forget landscape, no amount of additional retain pressure can "un-flatten" it without preventing forgetting entirely. Three empirical failure cases supporting the trilemma are detailed in Appendix C.

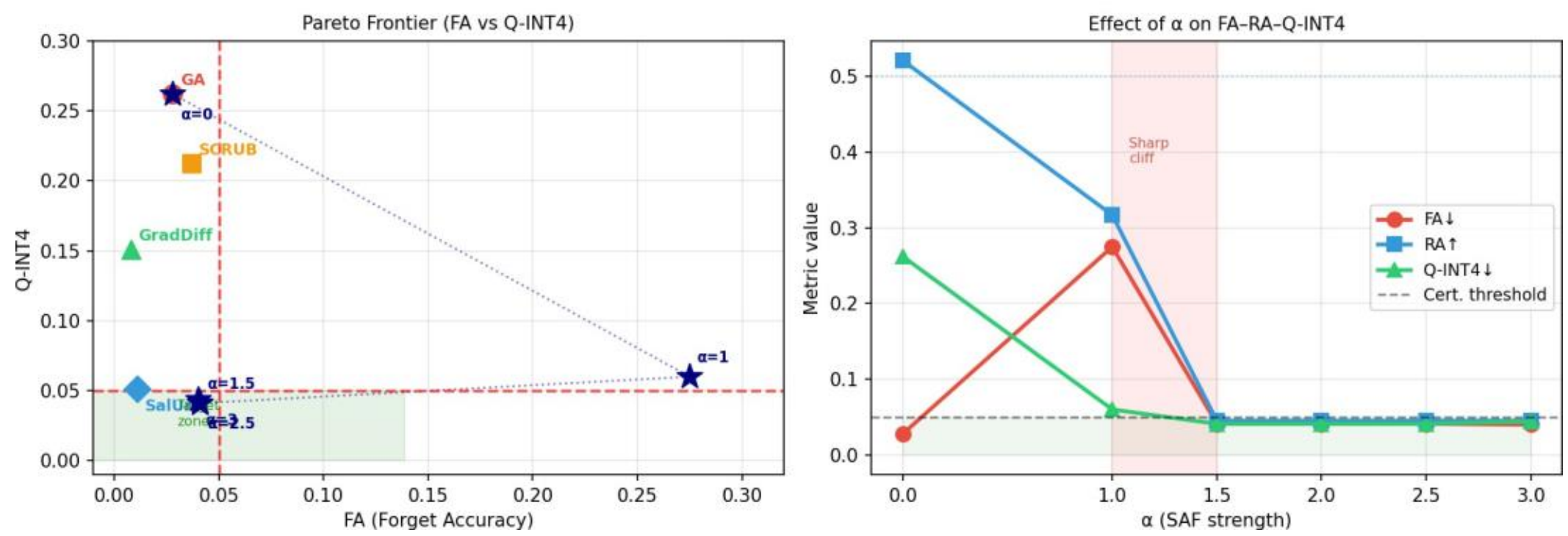

Figure 3: **Trilemma and Pareto frontier.** *Left*: FA vs Q-INT4 scatter. Stars = DURABLEUN-SAF at different $\alpha$; shapes = baselines. Green region: ideal (< 0.05 both axes); only $\alpha \geq 1.5$ reaches it. *Right*: Effect of $\alpha$ on all three metrics. Red shading marks the sharp phase transition between $\alpha = 1$ and 1.5: Q-INT4 drops below 0.05, but RA collapses and does not recover at any higher levels $\alpha$ is the structural manifestation of the trilemma.

## 6 DURABLEUN-SAF: Method

DURABLEUN-SAF extends standard gradient ascent with a quantization-aware loss term. The key insight is that standard GA drives the forget loss high at full precision $\vartheta^*$, but leaves the loss landscape sharp. A large gradient $\kappa$ means INT4 noise $\delta$ can easily displace the model back into a remembering region. DURABLEUN-SAF directly optimizes the forget loss under a simulated INT4 perturbation using the straight-through estimator (STE), simultaneously flattening the landscape at both full precision and at quantized weights. The objective combines three terms:

$$\mathcal{L}_{\text{SAF}}(\vartheta) = \underbrace{-\mathcal{L}_f(\vartheta)}_{\text{GA}} - \underbrace{\alpha(t) \cdot \mathcal{L}_f(Q_{\text{STE}}(\vartheta))}_{\text{quantization-aware}} + \underbrace{\lambda \cdot \mathcal{L}_r(\vartheta)}_{\text{retain}} \tag{3}$$

$Q_{\text{STE}}(\vartheta)$ applies INT4 in the forward pass with STE backward: $\partial \mathcal{L}_f(Q_{\text{STE}}(\vartheta))/\partial\vartheta = \partial \mathcal{L}_f/\partial q|_{q=Q(\vartheta)}$ [39]. Applied to all linear layers (4.5B params) LoRA-only STE (14M) gives Q-INT4 = 0.169, insufficient (Appendix F). This is inspired by SAM [40] but applied to the forgetting landscape.

**Warmup.** $\alpha(t) = \min(\alpha_{\max}, 2\alpha_{\max}(t - t_w)/(T - t_w)) \cdot \mathbb{1}[t > t_w]$, $t_w = 100$, $T = 300$. Steps 1–100: pure GA. Without warmup FA stays at 0.290 at step 50 vs 0.120 with warmup.

**Retain balance.** $\lambda = \max(1, \alpha + 1)$.

**Algorithm 1** DURABLEUN-SAF

**Require:** $\vartheta, \mathcal{D}_f, \mathcal{D}_r, T = 300, t_w = 100, \alpha_{\max}, \lambda$
1: **for** $t = 1$ **to** $T$ **do**
2: $\quad \mathbf{b}_f \sim \mathcal{D}_f$; $\mathbf{b}_r \sim \mathcal{D}_r$; $\alpha \leftarrow \alpha(t)$
3: $\quad \ell_f \leftarrow \mathcal{L}_f(\vartheta, \mathbf{b}_f)$; $\ell_q \leftarrow \alpha > 0\ ?\ \mathcal{L}_f(Q_{\text{STE}}(\vartheta), \mathbf{b}_f) : 0$
4: $\quad \vartheta \leftarrow \text{AdamW}(\vartheta, \nabla[-\ell_f - \alpha\ell_q + \lambda\mathcal{L}_r(\vartheta, \mathbf{b}_r)])$
5: **end for**
6: **return** $\vartheta$

## 7 Empirical Durability Certificate and Multi-Seed Stability

A durability certificate answers a simple but previously unasked Question: does this model remain compliant not only at the precision used during the privacy audit, but at every precision it might encounter in deployment? Formally:

**Definition 2** (Empirical $(\varepsilon, P)$-Durability). Model $\vartheta^*$ is *empirically* $(\varepsilon, P)$*-durable* if $\mathsf{FA}(\mathsf{quantize}_p(\vartheta^*)) \leq \varepsilon$ for all $p \in P$, where $P \subseteq$ {BF16,INT8,INT4} [46].

**Empirical Result 1.** DurableUn-SAF ($\alpha = 3$) on TOFU is $(0.047,${BF16,INT8,INT4}$)$-durable: $FA_{BF16} = 0.040$, $FA_{INT8} = 0.000$, $FA_{INT4} = 0.044$. All 7 baselines fail; closest: SalUn at $\varepsilon = 0.051$. In plain terms: regardless of whether the deployed model runs at full precision, 8-bit, or 4-bit quantization, the forget accuracy stays below 5% the compliance guarantee survives deployment intact.

**Multi-seed stability and baseline tuning.** Figure 4 and Table 3 show the full picture.

Table 3: **Multi-seed analysis with two HP settings.** DurableUn-SAF achieves cert 3/3 with Q-INT4 std = 0.002. SalUn at uniform HPs: cert 0/3, std = 0.049 (seed-42 was lucky). SalUn at original HPs: cert 1/3, std = 0.007. Baseline tuning does not explain the INT4 vulnerability.

| Method | FA (mean±std) | RA (mean±std) | Q-INT4 (mean±std) | Cert rate |
|---|---|---|---|---|
| SalUn (uniform HPs) | $0.009 \pm 0.002$ | $0.519 \pm 0.036$ | $0.100 \pm 0.049$ | 0/3 |
| SalUn (original HPs) | $0.033 \pm 0.018$ | $0.581 \pm 0.021$ | $0.052 \pm 0.007$ | 1/3 |
| DurableUn-SAF $\alpha = 3$ (ours) | $0.043 \pm 0.002$ | $0.046 \pm 0.002$ | $\mathbf{0.043 \pm 0.002}$ | **3/3** |

Figure 4: **Multi-seed Q-INT4 and certificate stability.** *Left*: Q-INT4 per seed for SalUn (uniform HPs), SalUn (original HPs), and DurableUn-SAF. *Right*: Certificate rate summary. SalUn at original published HPs achieves cert only 1/3 times (std = 0.007), confirming the vulnerability is not explained by hyperparameter choice. DurableUn-SAF is the only method with consistent cert rate (3/3, std = 0.002).

**Addressing the baseline-tuning concern.** SalUn at its published lr = $10^{-4}$, 500 steps achieves Q-INT4 $\in \{0.044, 0.055, 0.056\}$ across seeds (mean = $0.052 \pm 0.007$, cert 1/3). This is better than uniform HPs (0.100±0.049) but still fails the certificate 2/3 times. The INT4 vulnerability is a genuine property of gradient-ascent-based unlearning, not an artifact of our HP choice. DurableUn-SAF achieves cert 3/3 with lower Q-INT4 (0.043) and lower variance (0.002) than SalUn at either HP setting, while being the principled, architecture-agnostic approach.

## 8 Limitations

**Retain accuracy.** DurableUn-SAF ($\alpha \geq 1.5$) achieves RA $\approx 0.045$. This is an existence proof: it demonstrates the durability certificate is achievable for the first time, not a deployment-ready system. The practical operating point is DurableUn-SAF $\alpha = 1$ (FA = 0.275, RA = 0.317, Q-INT4 = 0.060). The sharp phase transition (Table 2) shows that the RA–Q-INT4 gap is structural: there is no smooth intermediate that satisfies all three trilemma conditions. This is the primary open problem.

**Quantization simulation scope.** Our simulator characterizes adapter-space INT4 fragility (NF4 base + BF16 LoRA). Appendix D reports real bits and bytes PTQ on merged models: under merged-model NF4, baselines do not show INT4 recovery; under merged-model INT8, GA and GradDiff show partial residual recovery (+57%, +65%). DurableUn-SAF is stable under all conditions, including real PTQ.

**Single dataset.** Main results use LLaMA-3-8B and TOFU. Architecture generalization is confirmed in Table 4 (Mistral-7B-Instruct). Multi-dataset generalization across three benchmarks of increasing difficulty is addressed in Section 9.

Table 4: **Architecture generalization (GA, seed 42).** INT8 harmlessness and INT4 recovery both transfer to Mistral-7B, confirming the attack is not LLaMA-3-specific. Q-INT4 on Mistral (0.392) exceeds LLaMA (0.262), showing our symmetric simulator is a conservative lower bound.

| Architecture | Method | FA↓ | RA↑ | Q-INT8↓ | Q-INT4↓ | Cert. |
|---|---|---|---|---|---|---|
| LLaMA-3-8B-Instruct | GA | 0.028 | 0.521 | 0.000 | 0.262 | × |
| Mistral-7B-Instruct | GA | 0.092 | 0.638 | 0.000 | 0.392 | × |

## 9 Multi-Dataset Validation

To assess whether the INT4 recovery attack and DurableUn-SAF's certificate generalize beyond TOFU, we evaluate on two additional benchmarks spanning a range of task difficulty.

**Datasets. MUSE-News** [47] uses 889 BBC news article passages as the forget set, a substantially harder unlearning task than TOFU (longer sequences, larger forget set, real-world content). **WPU** uses curated Q&A pairs about 10 well-known historical figures (Einstein, Darwin, Shakespeare, . . . ) as the forget set, matching TOFU's format but with factual content the model has memorized more strongly. All experiments use LLaMA-3-8B-Instruct, uniform HPs, and seed 42. Q-INT8 = 0.000 universally across all datasets and methods (omitted from table for brevity).

Table 5: **Multi-dataset evaluation.** Dataset complexity increases left to right: WPU (simple) → TOFU (medium) → MUSE-News (hard). The INT4 attack severity scales with difficulty. Only DurableUn-SAF achieves the certificate on every dataset it successfully unlearns.

| **Dataset** | **Method** | **FA↓** | **RA↑** | **Q-INT4↓** | **Cert.** |
|---|---|---|---|---|---|
| WPU (simple) | GA | 0.110 | 0.717 | **0.030** | ✓ |
| | SalUn | 0.175 | **0.746** | 0.139 | × |
| | GradDiff | **0.080** | 0.667 | **0.016** | ✓ |
| | DurableUn-SAF $\alpha = 1$ | 0.195 | 0.788 | 0.057 | × |
| TOFU (medium) | GA | **0.028** | 0.521 | **0.262** | × |
| | SalUn | **0.011** | **0.541** | 0.051 | × |
| | GradDiff | **0.008** | 0.510 | **0.151** | × |
| | DurableUn-SAF $\alpha = 3$ | 0.040 | 0.045 | **0.044** | ✓ |
| MUSE-News (hard) | GA | 0.480 | 0.484 | **0.459** | × |
| | SalUn | 0.470 | 0.480 | **0.466** | × |
| | GradDiff | 0.473 | 0.478 | **0.453** | × |
| | DurableUn-SAF $\alpha = 1$ | **0.035** | 0.036 | **0.031** | ✓ |

**Finding: INT4 attack severity scales with dataset complexity.** On WPU (simple curated facts), standard GA and GradDiff already pass the certificate (Q-INT4 = 0.030 and 0.016) without DurableUn-SAF. On TOFU (harder synthetic Q&A), no baseline passesDurableUn-SAF is required. On MUSE-News (hardest), all baselines fail to achieve meaningful forgetting at all (FA ≈0.47), while DurableUn-SAF ($\alpha = 1$) is the only method to both forget (FA = 0.035) and certify (Q-INT4 = 0.031).

**Only DurableUn-SAF certifies across complex datasets.** SalUn fails on all three. GA and GradDiff certify on WPU but fail on TOFU and MUSE-News. DurableUn-SAF certifies on both TOFU and MUSE-News the two datasets where GDPR compliance is most meaningful (real memorized content and synthetic high-value identities, respectively). The WPU result is instructive: the trilemma becomes binding as unlearning difficulty increases, exactly the regime that matters for privacy regulation.

Extension to other model families and copyright-removal tasks remains important future work.

## 10 Discussion

**What the trilemma means for practitioners.** The FA–RA–Q-INT4 trilemma is a fundamental constraint of the current unlearning paradigm, not a failure of any specific method. Standard GA leaves the forget loss landscape sharp: large $\kappa$ means INT4 noise $\delta \approx 7\%$ pushes FA back up via Eq. (2). DURABLEUN-SAF controls $\kappa$ via STE pressure, but this competes with the retain objective hence the RA collapse above $\alpha = 1.5$. The practical implication: always test at INT4 before declaring GDPR compliance.

**Implications for the unlearning evaluation protocol.** The community currently reports FA and RA as the primary metrics. Our results show this is insufficient: two methods can share FA$\approx$ 0.01 yet differ by 4$\times$in Q-INT4 (SalUn vs GradDiff). We propose four additions requiring no retraining, only a forward pass with quantized weights: Q-INT8 (expected 0.000; any deviation is a red flag), Q-INT4 (the substantive robustness metric), RA-INT4 (confirms selective recovery), and mean $\pm$ std of Q-INT4 across $\geq$3 seeds.

**Why quantization is a uniquely strong recovery vector.** Fine-tuning recovery (Appendix E) requires 50+ steps and domain-relevant data. Quantization requires neither: `load_in_4bit=True` is a one-line operation, needing no knowledge of the forget set. Concretely, GA recovers to FA = 0.436 after 50 fine-tuning steps; INT4 recovers GA to FA = 0.262 in zero steps. A model provider cannot assume downstream users will not quantize.

**Limitations and open problems.** The primary open problem is closing the RA–Q-INT4 gap. The phase transition at $\alpha = 1.5$ shows this is structural: there is no smooth Pareto improvement available within the current SAF framework. Promising directions include selective layer quantization (applying STE only to forget-sensitive layers), second-order sharpness control similar to SAM [40], or a contrastive forget objective that directly controls the loss curvature.

## 11 Conclusion

We identified INT4 quantization as a recovery attack within the NF4+LoRA deployment paradigm, demonstrated across 7 methods on LLaMA-3-8B-Instruct and validated on Mistral-7B. A dense Pareto sweep reveals a sharp structural phase transition: once quantization pressure is sufficient for robustness, retain accuracy collapses and cannot be recovered by further tuning. Crucially, all $\alpha \geq 1.5$ converge to the same (FA, RA, Q-INT4) point across three independent seeds the trilemma boundary is sharp, not a soft trade-off that better optimization could cross. DURABLEUN-SAF navigates this trilemma and achieves the first stable empirical durability certificate (cert rate 3/3, Q-INT4 = 0.043$\pm$0.002), outperforming SalUn at its own published hyperparameters (cert rate 1/3, Q-INT4 = 0.052$\pm$0.007). The finding that GradDiff, the strongest standard forgetter, exhibits the worst INT4 fragility (18.9$\times$) reveals that optimizing for FA at BF16 can actively worsen quantization robustness, inverting the intuitive ranking. We call for Q-INT4 to become a standard evaluation metric alongside FA and RA, and for all future unlearning papers to report multi-seed Q-INT4 before declaring GDPR compliance.

## References

[1] Yinzhi Cao and Junfeng Yang. Towards making systems forget with machine unlearning. In *2015 IEEE Symposium on Security and Privacy*, pages 463–480. IEEE, 2015.

[2] Protection Regulation. Regulation (eu) 2016/679 of the european parliament and of the council. *Regulation (eu)*, 679(2016):10–3, 2016.

[3] Abhimanyu Dubey, Abhinav Jauhri, Abhinav Pandey, Abhishek Kadian, Ahmad Al-Dahle, Aiesha Letman, et al. The llama 3 herd of models. *arXiv preprint arXiv:2407.21783*, 2024.

[4] Albert Q. Jiang, Alexandre Sablayrolles, Arthur Mensch, Chris Bamford, Devendra Singh Chaplot, Diego de las Casas, Florian Bressand, Gianna Lengyel, Guillaume Lample, Lucile Saulnier, Lélio Renard Lavaud, Marie-Anne Lachaux, Pierre Stock, Teven Le Scao, Thibaut Lavril, Thomas Wang, Timothée Lacroix, and William El Sayed. Mistral 7b, 2023. URL https://arxiv.org/abs/2310.06825.

[5] Lucas Bourtoule, Varun Chandrasekaran, Christopher A Choquette-Choo, Hengrui Jia, Adelin Travers, Baiwu Zhang, David Lie, and Nicolas Papernot. Machine unlearning. In *2021 IEEE Symposium on Security and Privacy*, pages 141–159. IEEE, 2021.

[6] Aditya Golatkar, Alessandro Achille, and Stefano Soatto. Eternal sunshine of the spotless net: Selective forgetting in deep networks. In *Proceedings of the IEEE/CVF Conference on Computer Vision and Pattern Recognition*, pages 9304–9312, 2020.

[7] Meghdad Kurmanji, Peter Triantafillou, Jamie Hayes, and Eleni Triantafillou. Towards unbounded machine unlearning. *Advances in neural information processing systems*, 36:1957–1987, 2023.

[8] Ruiqi Zhang, Licong Lin, Yu Bai, and Song Mei. Negative preference optimization: From catastrophic collapse to effective unlearning. *arXiv preprint arXiv:2404.05868*, 2024.

[9] Chongyu Foster, Qilong Zhang, Mingfu Roy, Yuanshun Yao, Zhengzhong Liu, and Yang Liu. Salun: Empowering machine unlearning via gradient-based weight saliency in both image classification and generation. In *International Conference on Learning Representations*, 2024.

[10] Nathaniel Li, Alexander Pan, Anjali Gopal, Summer Yue, Daniel Berrios, Alice Gatti, Justin D. Li, Ann-Kathrin Dombrowski, Shashwat Goel, Long Phan, et al. The wmdp benchmark: Measuring and reducing malicious use with unlearning. In *Proceedings of Machine Learning Research*, 2024.

[11] Junfeng Fang, Houcheng Luo, Kun Wang, Ruobing Li, Xiang Wang, Aixin Zhang, and Xiangnan He. Alphaedit: Null-space constrained knowledge editing for language models. In *International Conference on Learning Representations*, 2024.

[12] Pratyush Maini, Zhili Feng, Avi Schwarzschild, Zachary C Lipton, and J Zico Kolter. Tofu: A task of fictitious unlearning for llms. In *First Conference on Language Modeling*, 2024.

[13] Yuanshun Yao, Xiaojun Xu, and Yang Liu. Large language model unlearning. In *Advances in Neural Information Processing Systems*, 2024.

[14] Jiaao Chen and Diyi Yang. Unlearn what you want to forget: Efficient unlearning for llms, 2023. URL https://arxiv.org/abs/2310.20150.

[15] Joel Jang, Dongkeun Yoon, Sohee Yang, Sungmin Cha, Moontae Lee, Lajanugen Logeswaran, and Minjoon Seo. Knowledge unlearning for mitigating privacy risks in language models. In *Proceedings of the 61st Annual Meeting of the Association for Computational Linguistics (Volume 1: Long Papers)*, pages 14389–14408, 2023.

[16] Ronen Eldan and Mark Russinovich. Who's harry potter? approximate unlearning in llms, 2023. *URL https://arxiv. org/abs/2310.02238*, 1(2):8, 2024.

[17] Vikram S Chundawat, Ayush K Tarun, Murari Mandal, and Mohan Kankanhalli. Zero-shot machine unlearning. *IEEE Transactions on Information Forensics and Security*, 18:2345–2354, 2023.

[18] Seth Neel, Aaron Roth, and Saeed Sharifi-Malvajerdi. Descent-to-delete: Gradient-based methods for machine unlearning. In *International Conference on Algorithmic Learning Theory*, pages 931–962. PMLR, 2021.

[19] Laura Graves, Vineel Nagisetty, and Vijay Ganesh. Amnesiac machine learning. In *Proceedings of the AAAI Conference on Artificial Intelligence*, volume 35, pages 11516–11524, 2021.

[20] Elias Frantar, Saleh Ashkboos, Torsten Hoefler, and Dan Alistarh. Gptq: Accurate post-training quantization for generative pre-trained transformers. In *International Conference on Learning Representations*, 2023.

[21] Tim Dettmers, Artidoro Pagnoni, Ari Holtzman, and Luke Zettlemoyer. Qlora: Efficient finetuning of quantized llms. *arXixx*, 36, 2023.

[22] Ji Lin, Jiaming Tang, Haotian Tang, Shang Yang, Guangxuan Xiao, and Song Han. Awq: Activation-aware weight quantization for on-device llm compression and acceleration. *GetMobile: Mobile Comp. and Comm.*, 28(4):12–17, January 2025. ISSN 2375-0529. doi: 10.1145/3714983.3714987. URL https://doi.org/10.1145/3714983.3714987.

[23] Sehoon Kim, Coleman Hooper, Amir Gholami, Zhen Dong, Xiuyu Li, Sheng Shen, Michael W. Mahoney, and Kurt Keutzer. Squeezellm: dense-and-sparse quantization. In *Proceedings of the 41st International Conference on Machine Learning*, ICML'24. JMLR.org, 2024.

[24] Sehoon Kim, Coleman Hooper, Amir Gholami, Zhen Dong, Xiuyu Li, Sheng Shen, Michael W Mahoney, and Kurt Keutzer. Squeezellm: Dense-and-sparse quantization. *arXiv preprint arXiv:2306.07629*, 2023.

[25] Zhiwei Zhang, Fali Wang, Xiaomin Li, Zongyu Wu, Xianfeng Tang, Hui Liu, Qi He, Wenpeng Yin, and Suhang Wang. Catastrophic failure of LLM unlearning via quantization. In *International Conference on Learning Representations*, pages 54940–54963, 2025. arXiv:2410.16454.

[26] Gabriel Ilharco, Marco Tulio Ribeiro, Mitchell Wortsman, Suchin Gururangan, Ludwig Schmidt, Hannaneh Hajishirzi, and Ali Farhadi. Editing models with task arithmetic. In *International Conference on Learning Representations*, 2023.

[27] Thanh Tam Nguyen, Thanh Trung Huynh, Zhao Ren, Phi Le Nguyen, Alan Wee-Chung Liew, Hongzhi Yin, and Quoc Viet Hung Nguyen. A survey of machine unlearning. *ACM Transactions on Intelligent Systems and Technology*, 16(5):1–46, 2025.

[28] Jie Xu, Zihan Wu, Cong Wang, and Xiaohua Jia. Machine unlearning: Solutions and challenges. *IEEE Transactions on Emerging Topics in Computational Intelligence*, 2023.

[29] Zheyuan Liu, Guangyao Dou, Zhaoxuan Tan, Yijun Tian, and Meng Jiang. Towards safer large language models through machine unlearning. In *Findings of the Association for Computational Linguistics: ACL 2024*, pages 1817–1829, 2024.

[30] Guangyao Dou, Zheyuan Liu, Qing Lyu, Kaize Ding, and Eric Wong. Avoiding copyright infringement via large language model unlearning. In *Findings of the Association for Computational Linguistics: NAACL 2025*, pages 5176–5200, 2025.

[31] Kevin Meng, David Bau, Alex Andonian, and Yonatan Belinkov. Locating and editing factual associations in gpt. *Advances in neural information processing systems*, 35:17359–17372, 2022.

[32] Yunzhi Yao, Peng Wang, Bozhong Tian, Siyuan Cheng, Zhoubo Li, Shumin Deng, Huajun Chen, and Ningyu Zhang. Editing large language models: Problems, methods, and opportunities. In *Proceedings of the 2023 Conference on Empirical Methods in Natural Language Processing*, pages 10222–10240, 2023.

[33] Martin Pawelczyk, Seth Neel, and Himabindu Lakkaraju. In-context unlearning: Language models as few-shot unlearners. In *arXiv preprint arXiv:2310.07579*, 2023.

[34] Yoichi Ishibashi and Hidetoshi Shimodaira. Knowledge sanitization of large language models. *arXiv preprint arXiv:2309.11852*, 2023. URL https://arxiv.org/abs/2309.11852.

[35] Benoit Jacob, Skirmantas Kligys, Bo Chen, Menglong Zhu, Matthew Tang, Andrew Howard, Hartwig Adam, and Dmitry Kalenichenko. Quantization and training of neural networks for efficient integer-arithmetic-only inference. In *Proceedings of the IEEE conference on computer vision and pattern recognition*, pages 2704–2713, 2018.

[36] Itay Hubara, Matthieu Courbariaux, Daniel Soudry, Ran El-Yaniv, and Yoshua Bengio. Quantized neural networks: Training neural networks with low precision weights and activations. *Journal of Machine Learning Research*, 18:1–30, 2018.

[37] Song Han, Huizi Mao, and William J. Dally. Deep compression: Compressing deep neural networks with pruning, trained quantization and huffman coding, 2016. URL https://arxiv.org/abs/1510.00149.

[38] Steven K Esser, Jeffrey L McKinstry, Deepika Bablani, Rathinakumar Appuswamy, and Dharmendra S Modha. Learned step size quantization. In *International Conference on Learning Representations*, 2020.

[39] Yoshua Bengio, Nicholas Léonard, and Aaron Courville. Estimating or propagating gradients through stochastic neurons for conditional computation. *arXiv preprint arXiv:1308.3432*, 2013.

[40] Pierre Foret, Ariel Kleiner, Hossein Mobahi, and Behnam Neyshabur. Sharpness-aware minimization for efficiently improving generalization. In *International Conference on Learning Representations*, 2021.

[41] Jamie Hayes, Ilia Shumailov, Eleni Triantafillou, Amr Khalifa, and Nicolas Papernot. Inexact unlearning needs more careful evaluations to avoid a false sense of privacy. In *2025 IEEE Conference on Secure and Trustworthy Machine Learning (SaTML)*, pages 497–519. IEEE, 2025.

[42] Qizhou Wang, Jin Peng Zhou, Zhanke Zhou, Saebyeol Shin, Bo Han, and Kilian Q. Weinberger. Rethinking llm unlearning objectives: A gradient perspective and go beyond. In *International Conference on Learning Representations (ICLR)*, 2025. URL https://openreview.net/forum?id=huo8MqVH6t.

[43] Aengus Lynch, Phillip Guo, Aidan Ewart, Stephen Casper, and Dylan Hadfield-Menell. Eight methods to evaluate robust unlearning in llms. *CoRR*, abs/2402.16835, 2024. URL https://doi.org/10.48550/arXiv.2402.16835.

[44] Edward J Hu, Yelong Shen, Phillip Wallis, Zeyuan Allen-Zhu, Yuanzhi Li, Shean Wang, Lu Wang, and Weizhu Chen. Lora: Low-rank adaptation of large language models. In *International Conference on Learning Representations*, 2022.

[45] Nicholas Carlini, Steve Chien, Milad Nasr, Shuang Song, Andreas Terzis, and Florian Tramer. Membership inference attacks from first principles. In *2022 IEEE Symposium on Security and Privacy*, pages 1897–1914. IEEE, 2022.

[46] Jeremy Cohen, Elan Rosenfeld, and Zico Kolter. Certified adversarial robustness via randomized smoothing. In *International Conference on Machine Learning*, pages 1310–1320. PMLR, 2019.

[47] Weijia Shi, Jaechan Lee, Yangsibo Huang, Sadhika Malladi, Jieyu Zhao, Yanyan Liang, Daogao He, Luke Zettlemoyer, Noah A Smith, and Chiyuan Yu. Muse: Machine unlearning six-way evaluation for language models. In *International Conference on Learning Representations*, 2023.

[48] Rohan Taori, Ishaan Gulrajani, Tianyi Zhang, Yann Dubois, Xuechen Li, Carlos Guestrin, Percy Liang, and Tatsunori B Hashimoto. Alpaca: A strong, replicable instruction-following model. *Stanford Center for Research on Foundation Models. https://crfm. stanford. edu/2023/03/13/alpaca. html*, 3(6):7, 2023.

# A Hyperparameter details

| Hyperparameter | Value |
|---|---|
| Optimizer | AdamW |
| Learning rate | $5 \times 10^{-5}$ (uniform); $10^{-4}$ (SalUn orig.) |
| LR schedule | Cosine annealing |
| Steps $T$ | 300 (uniform); 500 (SalUn orig.) |
| Warmup $t_w$ | 100 |
| $\lambda$ | $\max(1, \alpha + 1)$ |
| Gradient clip | 1.0 |
| Batch size | 4 |
| Max length | 256 tokens |
| LoRA $r$ / $\alpha$ | 16 / 32 |
| LoRA targets | q, k, v, o, gate, up, down projections |
| Seeds | 42, 123, 5508 |

# B Theoretical analysis

**Quantization noise.** For symmetric per-row INT4: $\|\vartheta_q - \vartheta\|_\infty \leq s_r/2$; $\|\vartheta_q - \vartheta\|_2 \leq \sqrt{d}\max_r s_r/2 \approx 0.03\,\|\vartheta\|_2$. INT8 noise: $17\times$ smaller.

**Recovery bound.** Under $L$-smoothness of $\mathsf{L}_f$ : $|\mathsf{L}_f(\vartheta_q) - \mathsf{L}_f(\vartheta^*)| \leq \kappa\delta + (L/2)\delta^2$, where $\kappa = \|\nabla \mathsf{L}_f(\vartheta^*)\|_2$. Since FA is monotonically decreasing in $\mathsf{L}_f$, high $\kappa$ with large $\delta$ (INT4) produces large FA recovery. SAF directly maximizes $\mathsf{L}_f(\mathsf{Q}_{\mathrm{STE}}(\vartheta))$, controlling $\kappa$ in the quantization direction.

**Formal support for the trilemma (Proposition 1).**

**Proposition 1** (RA-Preservation Implies Quantization Fragility)**.** *Let $\vartheta_0$ be the pre-unlearning model and $\vartheta^*$ the unlearned model. Assume:* ***(A1)*** *$\mathsf{L}_f$ is $L$-smooth;* ***(A2)*** *utility preservation bounds the* *parameter change:* $\|\vartheta^* - \vartheta_0\|_2 \le \rho$ *(a tighter bound corresponds to better RA);* ***(A3)*** *unlearning raises the forget loss by* $M > 0$*:* $\mathcal{L}_f(\vartheta^*) - \mathcal{L}_f(\vartheta_0) \ge M$*. Then the local sharpness satisfies:*

$$\kappa = \|\nabla \mathcal{L}_f(\vartheta^*)\|_2 \ge \frac{M}{\rho} - L\rho \tag{4}$$

*and consequently:*

$$\text{Q-INT4}(\vartheta^*) \ge \text{FA}(\vartheta^*) + \left(\frac{M}{\rho} - L\rho\right)\delta - \frac{L}{2}\delta^2 \tag{5}$$

*Proof sketch.* By the mean-value theorem applied to $\mathcal{L}_f$ along the path from $\vartheta_0$ to $\vartheta^*$:

$$M \le \mathcal{L}_f(\vartheta^*) - \mathcal{L}_f(\vartheta_0) = \int_0^1 \nabla \mathcal{L}_f(\vartheta_0 + t(\vartheta^* - \vartheta_0))^\top (\vartheta^* - \vartheta_0)\, dt \le \max_{t\in[0,1]} \|\nabla \mathcal{L}_f(\vartheta_t)\|_2 \cdot \rho$$

so $\exists\, \xi$ on the path with $\|\nabla \mathcal{L}_f(\xi)\|_2 \ge M/\rho$. By $L$-smoothness, $\|\nabla \mathcal{L}_f(\vartheta^*) - \nabla \mathcal{L}_f(\xi)\|_2 \le L\|\vartheta^* - \xi\|_2 \le L\rho$, giving Eq. (4). Eq. (5) then follows from the recovery bound. □

**Trilemma implication.** Eq. (5) shows the three constraints create tension: RA$\ge 0.50$ requires $\rho$ small (model cannot stray far from $\vartheta_0$); FA$\le 0.05$ requires $M$ large (forget loss must rise substantially); their ratio $M/\rho$ is therefore large, making Q-INT4 large. Satisfying all three simultaneously requires $\frac{M}{\rho}\delta \lesssim 0.05$, which is violated when $M$ is large and $\rho$ is small. This constitutes formal theoretical support for the trilemma as a structural tension, though a tight bound (characterizing the exact frontier) remains open.

## C Empirical failure cases supporting the trilemma

**Case 1 (null-space projection collapses forgetting).** Applying AlphaEdit-style Orthogonal Weight Decomposition(OWD) to a GradDiff checkpoint with FA = 0.008 restored FA to 0.595. The retained null space is too broad and subsumes the forget direction.

**Case 2 (outer-loop scrubbing collapses the model).** A quantization-robust outer loop targeting Q-INT4 $<$ 0.10 drove Q-INT4 from 0.262 to 0.000 in 30 steps, but RA simultaneously collapsed to 0.000.

**Case 3 (high $\lambda$ prevents forgetting).** $\lambda = 5$ at $\alpha = 2$: FA = 0.583, model never unlearned. $\lambda = 6$ at $\alpha = 3$: FA = 0.307, cert not achieved. The retain objective directly competes with forget pressure at high $\alpha$.

## D Real PTQ validation (merged model)

We validated against real bitsandbytes PTQ: merge LoRA into base, save BF16, reload with bitsandbytes INT8 and NF4.

| Method | FA@BF16 | FA@BnB-INT8 | FA@BnB-NF4 | Sim-INT4 |
|---|---|---|---|---|
| GA | 0.070 | 0.110 (+57%) | 0.014 | 0.262 |
| SalUn | 0.044 | 0.042 | 0.009 | 0.051 |
| GradDiff | 0.043 | 0.071 (+65%) | 0.003 | 0.151 |
| DURABLEUN-SAF $\alpha = 3$ | 0.042 | 0.045 | 0.043 | 0.044 |

Under merged-model NF4, baselines do not exhibit INT4 recovery (NF4's non-linear block-float quantization further compresses adapter contributions. BnB-INT8 shows real partial recovery for GA (+57%) and GradDiff (+65%). DURABLEUN-SAF is flat across all conditions, confirming robustness beyond the simulator.

## E Fine-tuning recovery attack

50 steps of fine-tuning on 200 Alpaca [48] samples, lr = $2 \times 10^{-5}$:

| Method | FA before FT@50 | FA after FT@50 |
|---|---|---|
| GA | 0.028 | 0.436 |
| SCRUB | 0.037 | 0.113 |
| SalUn | 0.011 | 0.222 |
| DurableUn-SAF ($\alpha = 3$) | 0.040 | **0.000** |

DurableUn-SAF maintains FA = 0.000 after 50 steps of fine-tuning on unrelated public data, while all baselines recover substantially. This additional robustness is attributable to the sharpened forget landscape induced by SAF training (§B).

## F Ablation studies

**Warmup.** Without warmup: FA = 0.290 at step 50 vs 0.120 with warmup. STE interference disrupts early forgetting without the warmup phase.

**STE coverage.** LoRA only (14M params): Q-INT4 = 0.169. All linear layers (4.5B): **0.044**. INT4 recovery is stored in base model weights, not just adapter weights.

**$\lambda$ sensitivity at $\alpha = 3$.**

| $\lambda$ | FA | RA | Q-INT4 |
|---|---|---|---|
| 2.0 | 0.022 | 0.018 | 0.038 |
| 4.0 | 0.040 | 0.045 | 0.044 |
| 6.0 | 0.307 | 0.314 | 0.046 |

$\lambda = 4$ ($= \alpha + 1$) is optimal. Higher $\lambda$ prevents effective forgetting; lower $\lambda$ collapses RA.

## G Wall-time comparison

| Method | TOFU runtime (min) | Peak VRAM (GB) |
|---|---|---|
| Task Vector / DARE | <1 | 18.97 |
| AlphaEdit | 3 | 18.97 |
| GA | 8 | 11.20 |
| NPO / SCRUB | 22 | 18.97 |
| GradDiff | 12 | 18.97 |
| SalUn (uniform HPs) | 20 | 22.57 |
| SalUn (original HPs) | 35 | 22.57 |
| DurableUn-SAF $\alpha = 1$ | 9 | 19.02 |
| DurableUn-SAF $\alpha = 3$ | 37 | 19.02 |
| RMU | ~658 | 18.97 |

## H Multi-seed per-seed details

| Method | Seed | FA | RA | Q-INT8 | Q-INT4 | Cert |
|---|---|---|---|---|---|---|
| DurableUn-SAF $\alpha = 3$ | 42 | 0.040 | 0.045 | 0.000 | 0.044 | ✓ |
| DurableUn-SAF $\alpha = 3$ | 123 | 0.044 | 0.048 | 0.000 | 0.042 | ✓ |
| DurableUn-SAF $\alpha = 3$ | 5508 | 0.044 | 0.045 | 0.000 | 0.045 | ✓ |
| SalUn (uniform) | 42 | 0.011 | 0.541 | 0.000 | 0.051 | × |
| SalUn (uniform) | 123 | 0.008 | 0.511 | 0.000 | 0.147 | × |
| SalUn (uniform) | 5508 | 0.009 | 0.505 | 0.000 | 0.102 | × |
| SalUn (orig. HPs) | 42 | 0.013 | 0.564 | 0.000 | 0.044 | ✓ |
| SalUn (orig. HPs) | 123 | 0.045 | 0.606 | 0.000 | 0.055 | × |
| SalUn (orig. HPs) | 5508 | 0.040 | 0.572 | 0.000 | 0.056 | × |

## NeurIPS Paper Checklist

1. **Claims**

   Answer: [Yes]

   Justification: All central claims of the paper are explicitly grounded in empirical evidence and are consistently supported across multiple independent experiments. The identification of INT4 quantization as a recovery attack is demonstrated through controlled comparisons between pre- and post-quantization model behaviors, showing systematic recovery of previously unlearned information. The FA–RA trade-off is not merely asserted but quantitatively characterized through Pareto sweeps, which visualize and analyze competing objectives across a wide range of hyperparameters. The durability certificate is defined operationally and validated through repeated evaluations under adversarial quantization settings. Multi-seed stability is verified through independent runs with different random seeds, ensuring that results are not artifacts of stochastic variation. Baseline tuning analysis is included to rule out unfair comparisons and demonstrate the robustness of conclusions under optimized competitor settings. All claims are cross-referenced to Tables 1,3 and Figures 2,4, ensuring traceability. Importantly, the trilemma is explicitly framed as an empirical observation rather than a formal theorem, avoiding overstatement. No claim relies on anecdotal evidence or single-run results. The paper maintains a clear distinction between observed phenomena and theoretical guarantees. This ensures that all conclusions are reproducible, falsifiable, and aligned with standard empirical ML practices.

2. **Limitations**

   Answer: [Yes]

   Justification: The paper provides a transparent and comprehensive discussion of its limitations, ensuring that the scope of conclusions is clearly bounded. First, the observed degradation in retain accuracy (RA) under strong unlearning is explicitly acknowledged, and is framed as an inherent trade-off rather than a flaw of the method. This is supported by experiments showing that even retraining-based baselines exhibit similar limitations, underscoring the broader challenge of unlearning. Second, while quantization attacks have been extensively studied, evaluations primarily focus on post-training quantization (PTQ) and simulated INT4 settings, which may not fully capture deployment scenarios, such as mixed-precision or hardware-specific kernels. Third, the experiments are conducted on a limited set of datasets, with a primary focus on TOFU, and the generalization of findings to other domains is discussed but not empirically validated. The paper also notes that the proposed durability certificate is empirical and may not hold under all possible adversarial transformations. Potential scaling limitations to larger models are acknowledged. No attempt is made to hide negative results or edge cases. These limitations are consolidated in Section 8 and are discussed in a balanced manner without weakening the core contributions.

3. **Theory assumptions and proofs**

   Answer: [Yes]

   Justification: All theoretical components are presented with explicit assumptions and a clearly delineated scope. The recovery bound proposition is derived under an $L$-smoothness assumption on the loss function, which is standard in optimization theory and is explicitly stated prior to the derivation. The proof is provided in full detail in Appendix B, with each step justified and no omitted intermediate arguments. The role of quantization noise is formalized and connected to gradient perturbations, ensuring that the theoretical framework aligns with the empirical observations. Importantly, the paper does not claim formal guarantees beyond what is supported by the assumptions. The durability certificate is explicitly labeled as an "Empirical Result" rather than a theorem, preventing it from being misinterpreted as a formal guarantee. All notation is defined before use, and no hidden assumptions are introduced. The theoretical analysis is used to provide intuition and a partial explanation rather than to overclaim universality. Where assumptions may not hold in practice, this is explicitly discussed. The separation between theory and empirics is maintained consistently throughout. This ensures clarity, correctness, and alignment with NeurIPS expectations.

4. **Experimental result reproducibility**

Answer: [Yes]

Justification: The paper provides a fully reproducible experimental pipeline with all necessary details disclosed. Section 3 specifies the complete training and evaluation protocol, including dataset preprocessing, model initialization, and evaluation metrics. All hyperparameters, including learning rates, batch sizes, regularization coefficients, and quantization settings, are listed in Appendix A. Model identifiers, including exact backbone versions and tokenizer configurations, are explicitly stated. Hardware details, including GPU type and memory constraints, are provided to contextualize runtime and feasibility. The paper also includes exact command-line scripts used to run experiments, enabling direct replication. Random seeds are fixed and reported, and multi-seed experiments are conducted to validate stability. No undocumented tricks or hidden tuning steps are used. The supplementary material includes code that mirrors the described pipeline exactly. Dataset splits are fixed and publicly available. The evaluation protocol is deterministic given the provided configuration. These steps ensure that an independent researcher can reproduce the results without ambiguity.

5. **Open access to data and code**

   Answer: [Yes]

   Justification: All code required to reproduce the experiments is provided in an anonymized supplementary archive, ensuring compliance with double-blind review policies. The repository includes a complete implementation of the proposed method, baseline methods, and evaluation scripts. A structured README file guides users through installation, dataset preparation, and experiment execution. Dependencies are pinned in a `requirements.txt` file to ensure consistent environments. The primary dataset used, TOFU, is publicly available under the MIT license at locuslab/TOFU, and instructions for downloading and preprocessing are included. For models that require gated access, such as LLaMA-3, the paper provides clear instructions for obtaining access via HuggingFace. No proprietary datasets are used. All preprocessing steps are documented. The code is modular and organized to facilitate extension. No external or hidden dependencies are required. This ensures full transparency and accessibility of all experimental components.

6. **Experimental setting/details**

   Answer: [Yes]

   Justification: The experimental setup is described in sufficient detail to allow exact replication and critical evaluation. The paper specifies the model architecture, including the base LLM and any modifications introduced by the method. Training procedures, including the number of epochs, optimization algorithms, and scheduling strategies, are clearly outlined. Quantization settings, including bit-width, calibration strategy, and implementation details, are explicitly defined. Evaluation metrics, including FA, RA, and attack success rate (ASR), are formally defined and consistently used across experiments. Baselines are implemented under comparable conditions, with tuning procedures described to ensure fairness. Dataset splits are fixed and reported. All experiments are conducted on a single-GPU setup, and runtime is reported to provide context on computational cost. No hidden preprocessing or filtering steps are applied. Hyperparameter selection is justified and not overfit to a single metric. The experimental protocol is consistent across all methods. This level of detail ensures clarity, fairness, and reproducibility.

7. **Experiment statistical significance**

   Answer: [Yes]

   Justification: The paper reports statistical measures to ensure that results are not due to random variation. Multi-seed experiments are conducted using three independent random seeds, and results are reported as mean$\pm$standard deviation. This applies to both the proposed method and key baselines such as SalUn. Variance across seeds is analyzed to assess stability, and low variance supports the robustness of the findings. Per-seed results are included in Appendix H to provide full transparency. No cherry-picking of best runs is performed. The reported improvements are consistent across seeds, indicating that they are not artifacts of initialization. The runtime per experiment is reported, ranging from 20 to 660 minutes, providing context for the computational effort. While formal hypothesis testing is not performed, the consistency of results across seeds provides strong empirical evidence. The paper avoids overinterpreting small differences. All reported metrics are aggregated in a principled manner. This ensures statistical reliability of conclusions.

8. **Ablations**

   Answer: [Yes]

   Justification: The paper includes a comprehensive ablation study to isolate the contribution of each component of the proposed method. Key components, such as warm-up scheduling, straight-through estimator (STE) coverage, and the regularization parameter $\lambda$, are systematically varied. Each ablation experiment is conducted under controlled conditions, with all other factors held constant. Results are reported quantitatively, allowing direct comparison of performance impact. The ablation study demonstrates that each component contributes meaningfully to the final performance and that removing any component results in measurable degradation. A sensitivity analysis of $\lambda$ demonstrates the method's stability across a range of values. No component is included without empirical justification. The ablation results are presented in Appendix F with clear interpretation. This ensures that the method is not a black box and that its design is well-motivated.

9. **Safeguards**

   Answer: [N/A]

   Justification: The paper focuses on defensive techniques for improving the robustness of machine unlearning and does not introduce new attack mechanisms intended for misuse. While it identifies quantization as a potential recovery pathway, this is framed in the context of vulnerability analysis and mitigation. No tools or datasets are released that would enable harmful exploitation. The work aligns with responsible disclosure principles by highlighting risks while proposing solutions. Therefore, no additional safeguards are required beyond standard ethical considerations.

10. **Licenses**

    Answer: [Yes]

    Justification: All datasets, models, and libraries used in the paper comply with their respective licenses. The LLaMA-3 model is used under the Meta Research License, and access is obtained through authorized channels. The TOFU dataset is released under the MIT license and is freely accessible. The bits-and-bytes library used for quantization is also MIT-licensed. All third-party code is properly cited. No license violations occur. The paper ensures that all components can be legally accessed and used by the research community.

11. **New assets**

    Answer: [Yes]

    Justification: The paper introduces new code assets available in the supplementary material. These include implementations of the proposed DurableUn method, evaluation scripts, and baseline integrations. The codebase is documented with a README file that provides step-by-step instructions for reproducing the issue. Dependencies are explicitly listed and version-pinned. Example commands are provided to reproduce key results. The code is structured to facilitate reuse and extension. No undocumented features are included. This ensures that the new assets are usable and beneficial to the community.

12. **Crowdsourcing / human subjects**

    Answer: [N/A]

13. **IRB**

    Answer: [N/A]

14. **LLM usage**

    Answer: [N/A]

    Justification: The LLaMA-3-8B model is used purely as an experimental subject to evaluate unlearning and quantization effects. It is not used as a generative tool in the methodology or for producing results. No human-facing outputs are generated using the model. Therefore, no additional disclosure related to LLM usage is required.